\documentclass[journal]{IEEEtran}
\usepackage{color,array}
\usepackage[english]{babel}
\usepackage{inputenc}
\usepackage{algorithm}
\usepackage{algorithmicx}
\usepackage{algcompatible}
\usepackage{algpseudocode}
\usepackage{makecell}
\usepackage{cite}
\usepackage{pbox}
\usepackage{amsmath,amssymb}

\usepackage[pdftex]{graphicx}
\usepackage[figurename=Fig.,font=small,skip=2 pt]{caption}
\graphicspath{{../pdf/}{../jpeg/}}
\graphicspath{ {./img/}}
\graphicspath{ {./img1/}}
\DeclareGraphicsExtensions{.pdf,.jpeg,.png}
\usepackage{epstopdf}
\usepackage{multirow}
\usepackage{accents}
\newcommand{\justified}{%
  \rightskip\z@skip%
  \leftskip\z@skip}
\usepackage{multicol}
\usepackage{enumerate}
\usepackage{array}
\usepackage{color,soul}
\usepackage{pbox}

\usepackage{comment}

\usepackage{stfloats}
\usepackage{amssymb}

\usepackage{setspace}
\usepackage{amsmath}
\usepackage{float}
\usepackage[table]{xcolor}
\definecolor{Gray}{gray}{0.85}
\usepackage{verbatim} 
\usepackage{color, colortbl}
\definecolor{Gray}{gray}{0.92}

\def\l{\left}   
\def\r{\right}   

\include{macros}

\begin{document}
\title{Guided Sampling-based Evolutionary Deep Neural Network for Intelligent Fault Diagnosis}
\author{Arun K. Sharma,~\IEEEmembership{Student Member,~IEEE and}
        Nishchal K. Verma,~\IEEEmembership{Senior Member,~IEEE}
      
\thanks{Arun K. Sharma and Nishchal K. Verma are with the Dept. of Electrical Engineering, Indian Institute of Technology, Kanpur, India.
e-mail: arnksh@iitk.ac.in and nishchal@iitk.ac.in}
}

\maketitle

\begin{abstract} The diagnostic performance of most of the deep learning models is greatly affected by the selection of model architecture and hyperparameters. Manual selection of model architecture is not feasible as training and evaluating the different architectures of deep learning models is a time-consuming process. Therefore, we have proposed a novel framework of evolutionary deep neural network which uses policy gradient to guide the evolution of DNN architecture towards maximum diagnostic accuracy.  We have formulated a policy gradient-based controller which generates an action to sample the new model architecture at every generation such that the optimality is obtained quickly. The fitness of the best model obtained is used as a reward to update the policy parameters. Also, the best model obtained is transferred to the next generation for quick model evaluation in the NSGA-II evolutionary framework. Thus, the algorithm gets the benefits of fast non-dominated sorting as well as quick model evaluation. The effectiveness of the proposed framework has been validated on three datasets: the Air Compressor dataset, Case Western Reserve University dataset, and Paderborn university dataset.
\end{abstract}  



\maketitle

\IEEEdisplaynontitleabstractindextext

\IEEEpeerreviewmaketitle

\ifCLASSOPTIONcompsoc
\IEEEraisesectionheading{\section{Introduction}\label{sec:introduction}}
\else
\section{Introduction}
\label{sec:introduction}
\fi
\IEEEPARstart {W}{ith} the advancement in modern computational technology, machine learning-based intelligent fault diagnosis has become an integral part of almost all industrial sectors. Intelligent fault diagnosis refers to the preventive maintenance of industrial machines using machine learning-based data analysis and fault class detection \cite{snandi, asiddi, rn1, chenX, rfIFD, aks2, sparseAE}. Intelligent fault diagnosis of industrial machines faces many challenges: (i) unavailability or limited availability of labeled data as running the machine in a faulty state with real-time load is an uneconomical task or sometimes not possible, (ii) training the deep learning algorithms with limited availability of labeled dataset faces problem of overfitting, (iii) selection of best model architecture suitable for the given dataset. Several methodologies have been introduced for cross-domain intelligent fault diagnosis with unavailability or limited labeled data availability. For example, domain-adversarial training of neural networks \cite{dann_jmlr, wLu}, cross-domain fault diagnosis under limited availability of labeled dataset \cite{cross-domain, dctln, aks_quick}. A quick learning mechanism \cite{aks_quick} solves the problem in training the deep learning model with limited availability of labeled target samples by using net2net transformation followed by domain adaptation-based fine-tuning. However, the performance of all these methods fault diagnosis is greatly affected by the selection of deep neural network (DNN) model architecture .

Therefore, our main objective is to investigate and develop an algorithm that can find the best suitable architecture for the fault diagnosis with the dataset under variable operating conditions of the machines. Although the hyper-parameter optimization is orthogonal to the model training, it greatly affects the diagnostic performance of the model.  Therefore, research on hyper-parameter optimization or neural architecture search (NAS) has gained very much attention \cite{tl-gdbn, autoML, Loghmanian2012, Wang2019, sun_pso, Ysun2020, TL_aks}. The NAS methods are mainly categorized as: (i) random search and grid search \cite{BJ}, (ii) surrogate model-based optimization \cite{LIO}, (iii) reinforcement learning \cite{BB_nn_rein}, (iv) genetic algorithm \cite{Loghmanian2012, Wang2019, sun_pso, Ysun2020, TL_aks}, (v) gradient descent \cite{darts}, and (vi) hybrid algorithms \cite{yang_cars}. In the aforementioned methods, the biggest challenge for architecture search is the model evaluation because of the complex training mechanism for most of the DNN models \cite{autoML}. In genetic algorithm-based NAS methods \cite{Loghmanian2012, Wang2019, sun_pso, Ysun2020, TL_aks}, the hyper-parameters of model to be optimized are encoded as an individual (chromosomes). A set of such individuals (population) are evaluated at each generation to evolve and find the best model. The effectiveness of the evolution process depends on (i) fitness evaluation strategy, (ii) sorting, and (iii) crossover and mutation strategy. The fitness evaluation involves the training and testing of the individuals (models) for the given dataset which is a time-consuming process for deep neural networks. For example, regularized evolution of image classifier \cite{regEvoImC} with 450 K40 GPU takes 3150 GPU days. Therefore, fault diagnosis with these NAS methods becomes infeasible as most of the industrial applications require a faster mechanism to quickly search and train a suitable architecture of DNN.  

Architecture optimization using fast non-dominated sorting genetic algorithm II (NSGA-II \cite{nsga}) get benefits of faster sorting method and therefore faster evolution \cite{nsgaNetV1}. However, this method too requires training and testing of individuals at each generation from scratch and therefore becomes a time-consuming process for fault diagnosis applications. EvoN2N \cite{TL_aks} uses the concept of knowledge transfer for fitness evaluation in the NSGA-II based framework. The quick fitness evaluation with fast NSGA-II makes the algorithm faster compared to the state-of-the-art evolutionary NAS methods. This method uses crossover and mutation-based exploration and exploitation to find the best DNN architecture in the given search space. Similar to conventional genetic algorithms, this process requires the large number of generations to converge. To address the issue of slow convergence in evolutionary NAS methods, we introduce a guided sampling-based evolution that makes the convergence faster while exploiting the search space with the help of a reward-based controller instead of conventional crossover and mutation. The key contributions of this work are highlighted below
\begin{enumerate}[i)]
    \item We have formulated the \textbf{guided sampling-based evolution of DNN architecture} (\textbf{GS-EvoN2N}) using policy gradient.
    \item We have introduced a novel method of \textbf{mean-variance-based mutation} to exploit the search space to obtain the optimality.
    \item We have formulated a \textbf{mean-variance update law using policy gradient} to guide the sampling of the DNN architecture to reach the optimality faster.
    \item We have adopted the quick model revaluation strategy based on the knowledge transfer mechanism by transferring the knowledge of the best model obtained at every generation.
\end{enumerate}

The rest of the article is organized as follows. Table \ref{tab:symbols} summarizes the symbols commonly used throughout the literature except the index variables on summation and the loop variables in algorithmic steps. Section \ref{prob_formulation} defines the objective problem. Section \ref{sec:relatedWorks} briefly discusses the related works and the theoretical backgrounds. Section \ref{sec:proposed} explains the implementation details of the proposed framework of GS-EvoN2N. Section \ref{expD} discussed the effectiveness of the proposed framework for fault diagnosis under various load and operating conditions of the machine. And finally, Section \ref{conclusion} concludes the whole paper.

\begin{table}[!ht]
\centering %
\caption{\textsc{List of Symbols}} %
\label{tab:symbols}
\resizebox{0.35\textheight}{!}{%
\begin{tabular}{c|c}
\hline\hline
{Symbol}  &  Description \\
\hline\hline
 $\mathcal{D}^{tr}$, $\mathcal{D}^{val}$, \& $\mathcal{D}^{te}$  &  Dataset, training, validation \& test dataset \\
\hline
 $\textbf{X}\in\Re^{(n_s\times n_f)}$, $\textbf{y}\in\Re^{n_s}$  &  Input data, Output labels\\
\hline
$n_s$ \& $n_f$ & Number of samples \& features\\
\hline
$C$      & number class of the dataset\\
\hline
$\Psi_t$ & DNN model at generation $t$\\
\hline
$\Psi_t^{\dagger}$ & Best DNN model at generation $t$\\
\hline
$\Lambda_t,\,\mathcal{R}_t$ & Fitness matrix, Rank at generation $t$\\
\hline
 $P_t,\, Q_t$ & Population, Offspring at generation $t$\\
 \hline
 $n_p$ & number hidden layers in $p^{th}$ model\\
\hline
$h_k$    & Number nodes in $k^{th}$ hidden layer\\
\hline
\end{tabular}}
\end{table}

\section{Problem Statement}\label{prob_formulation}
Let the training dataset, validation dataset, and test dataset are $\mathcal{D}^{tr}=\l(\textbf{X}^{tr}, \textbf{y}^{tr}\r)$, $\mathcal{D}^{val}=\l(\textbf{X}^{val}, \textbf{y}^{val}\r)$, and $\mathcal{D}^{te}=\l(\textbf{X}^{te}, \textbf{y}^{te}\r)$, respectively where $\textbf{X}\in\Re^{(n_s\times n_f)}$ be the input data with $n_s$ samples \& $n_f$ features and $\textbf{y}\in\Re^{n_s}$ be the corresponding output label. The objective of optimal DNN architecture search for fault classification is mathematically be formulated as 
\begin{flalign}
   & \Psi^{\dagger} =  \mathcal{H}\l(P,\; \mathcal{D}^{tr}, \;\mathcal{D}^{val}\r) \\
  &  \hat{y}^{te} \; = \; \mathcal{F}\l(\Psi^{\dagger},\; X^{te}\r) 
\end{flalign}
where $\mathcal{H}(.)$ denotes the optimization function to get the best model $\Psi^{\dagger}$ with optimal parameters for the training dataset and $\mathcal{F}(.)$ is the feed-forward DNN function which predicts the fault class $\hat{\textbf{y}}^{te}$ for the test data $\textbf{X}^{te}$.

\section{Related Works and Theoretical Background}\label{sec:relatedWorks}
\subsection{Deep Neural Network (DNN)}
\begin{figure}[!ht]
\centering
\includegraphics[width=8.5cm]{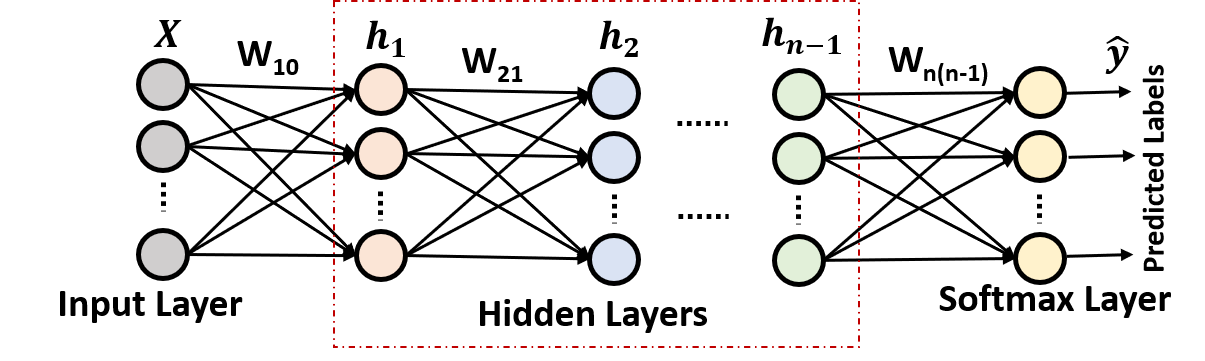}
\caption{SAE with softmax classifier (DNN model: $\Psi$)}
\label{fig:SAE}
\end{figure}

The deep neural network (DNN): a multi-layered neural network is the most popular technique for pattern recognition via non-linear feature transformation in multiple stages \cite{hinton}. From the training point of view, DNN can be considered as two parts: Stack of a given number of auto-encoders ( also called stacked auto-encoder: SAE) \cite{bengio} and a classifier usually softmax classifier as output layer. First, a greedy layer unsupervised training is used to train each of the auto-encoder (AE) in the SAE. Then, the SAE stacked with the classifier at the end layer is fine-tuned using a labeled training dataset. The SAE with softmax classifier (DNN model $\Psi$) is depicted in Fig. \ref{fig:SAE}.

\subsection{Intelligent Fault Diagnosis} Recently, with the advent of advanced machine learning techniques and the availability of fast computational resources, the data-driven intelligent fault diagnosis method has gained much popularity, \cite{snandi, asiddi, chenX, rn1}. In these methods, various machine learning techniques are utilized to learn the specific signature of the recorded signals like current, vibration, temperature, etc., and thereafter identify the existence of machinery fault using the test samples. Neural network (NN) \cite{nn1}, Support vector machine (SVM) \cite{svm, svm1}, and random forest (RF) classifier \cite{rfIFD} have been very effectively used for intelligent fault diagnosis and have been proved to be the baseline method for pattern recognition. But the diagnostic performances by these methods are reduced due to high sparsity and low-quality features in the dataset \cite{c1}. The intelligent fault diagnosis using deep learning methods has gained much attention due to its capability of multi-scale hierarchical feature transformation and large dimensional data handling, \cite{sparseAE, rzhao, dctln}. However, using a deep neural network for fault diagnosis faces a major challenge of training from scratch for every new operating condition of the machines. The recent trend of using deep transfer learning methods for domain adaptation has been very effective for fault diagnosis under changeable operating conditions \cite{sjPan2011, longRTRL, wLu, lwen, cross-domain, weigh_dom, aks_quick}. However, the diagnosis performance by these methods is very much affected by the selection of the architecture of the deep neural network and the other hyper-parameters.

\subsection{Neural Architecture Search (NAS)} The main objective of NAS methods is to obtain the optimal architecture in a given search space with the best model performance \cite{autoML}. There are three important aspects of the NAS methods: (i) formulation of the search space, (ii) the architecture optimizer, and (iii) the model evaluation. Formulation of search space defines the design format for the model architecture. It can be categorized into four groups: (i) cell-based (ii) entire-structured, (iii) morphism-based, and (iv) hierarchical search space. The most important aspect of NAS methods is the model evaluation as it is computationally very expensive to train each model during the search process and evaluate on the unseen dataset. To accelerate the evolution, various mechanism have been suggested for the model evaluation \cite{autoML, regEvoImC,  cvpr, sharpdarts, LIO, NIPS, NIPS2011, paraShare}. K. Kandasamy, \textit{et.al}, \cite{NIPS} suggested learning the curve extrapolation for the model performance evaluation instead to train and evaluate the actual architecture. H. Pham \textit{et. al}, \cite{paraShare} proposed the method of parameter sharing for the faster training and evaluation of the model architecture. 

Another important aspect of NAS methods is the architecture optimizer (AO). 
The objective automatic AO is to automatically guide the model architecture search in a direction to get the best suitable model for a given dataset. The AO methods adopted by various researchers can be categorized as (i) random search (RS)  (ii) grid search (GS), (iii) surrogate model-based optimization (SMBO), (iv) gradient descent (GD), (v) reinforcement learning (RL), (vi) genetic algorithms (GA), and (vii) hybrid methods. In the RS method, the search optimizer tries different architecture randomly from the defined search space, \cite{BJ}, whereas, in GS, the search method uses a grid to sample and evaluate the model architecture \cite{sharpdarts}. SMBO methods use Basian optimization \cite{LIO, NIPS, NIPS2011} or neural networks \cite{smbo_nn} as a surrogate model of the objective function to obtain the most promising solution (the model architecture). Gradient descent-based method uses softmax function to find the optimal architecture over a continuous and differentiable search space \cite{darts}. In RL-based NAS \cite{BB_nn_rein, zophRL}, a controller (usually, a recurrent neural network) generate an action to sample a new architecture. The observation (state) \& the reward from the environment is used to update the controller policy to generate new architecture samples. Here, the training \& validation process of the sampled neural network is treated as the environment that returns back the validation accuracy. GA-based NAS \cite{Loghmanian2012, Wang2019, sun_pso, Ysun2020, DE_PG, nsgaNetV1}, use heuristic search to find the best performing architecture over a given search space. In these methods, heuristically sampled neural architectures are trained and evaluated using the convention neural training methods and the performance metrics are used as fitness for evolution to obtain the optimal architecture. The main challenge of these methods is the fitness evaluation of the individual model All of these methods of AO have their own merits and demerits. The hybridization of two of the above methods may give a significant improvement in the search efficiency, called the hybrid method of AO \cite{renas, yang_cars, surrogateEA}.

\subsection{Policy Gradient} Policy gradient (PG) is a tool to optimize the controller policy for reinforcement learning algorithm \cite{PG1, PG2}. The controller policy is the parameterized function that defines the learning agent's way to act on the environment to get maximum reward Fig. \ref{Fig: flow diagram}. The reward defines the good or bad effect of the action taken by the policy towards the fulfillment of the optimal objective. 
Let the parameter vector be $\theta_t$, then the parameterized function policy is represented as $\pi_{\theta_t}(a_t|s_t)$, where $a_t$ and $s_t$ represent action and state at given time $t$. Let the action $a_t$ produces reward $r_{t+1}$ from the environment, then the trajectory of state, action and reward can be represented as $\l((s_0, a_0, r_1), (s_1, a_1, r_2), ....(s_t, a_t, r_{t+1})\r)$. The policy parameter $\theta_t$ can be updated using policy gradient as
\begin{equation}\label{eq:PG}
    \theta_{t+1} = \theta_t + \eta_t\nabla_{\theta_t}\mathcal{J}(\theta_t)
\end{equation}
where $\eta_t$ denotes the learning rate at time $t$, usually a constant real number. $\nabla_{\theta_t}\mathcal{J}(\theta_t)$ denotes the policy gradient and can be calculated using expected of cumulative reward ${U}_t$ over the time $t$ as follows.
\begin{align}
    \nabla_{\theta_t}\mathcal{J}(\theta_t) = \nabla_{\theta_t}E\l[{U}_t\r] = \nabla_{\theta_t}\int_t \pi(\tau)r(\tau)d(\tau)\\
   = E\l[r(\tau).\nabla_{\theta_t}\log\pi_{\theta_t}(\tau)\r]
\end{align}
\begin{equation}
\resizebox{0.33\textheight}{!}{%
    $\nabla_{\theta_t}\mathcal{J}(\theta_t) =\frac{1}{N}\sum_{k=1}^{N}r(k)\l(\sum_{t=1}^{T}\nabla_{\theta_t}\log\pi_{\theta_t}\l(a_t|a_{t-1:1};\theta_t\r)\r)$}\label{eq:PGd}
\end{equation}

\section{Proposed Framework}\label{sec:proposed} 
\begin{figure}[!ht]
\centering
\includegraphics[width=8.0cm]{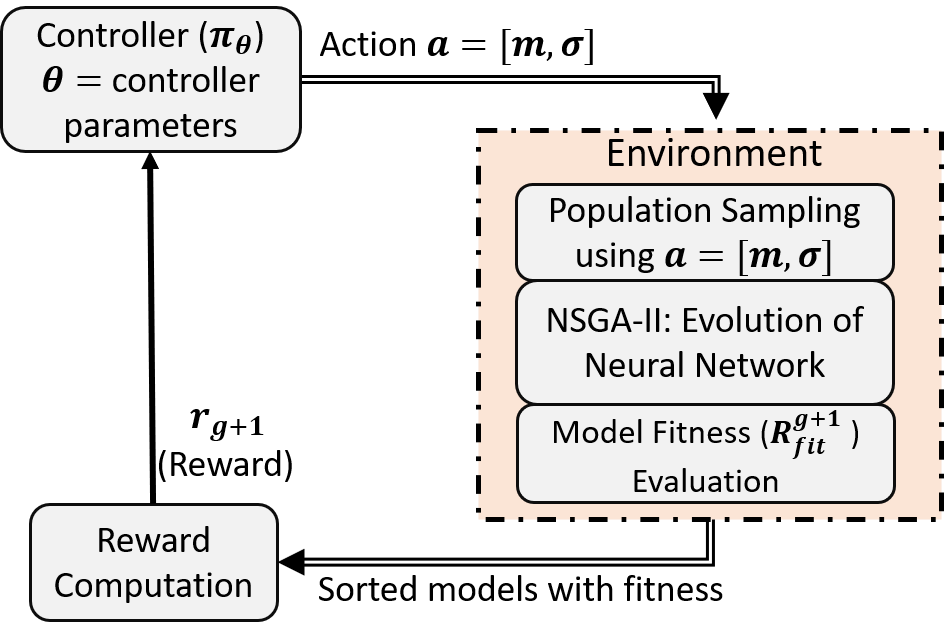}
\caption{Flow Diagram of guided sampling based NSGA-II.}
\label{Fig: flow diagram}
\end{figure}

In this section, the proposed framework of guided sampling-based evolutionary DNN (GS-EvoN2N) is described in detail. The Fig \ref{Fig: flow diagram} shows the schematic of the workflow  of GS-EvoN2N. In the figure, DNN architecture optimization in the NSGA-II framework constitute the environment. The fitness of the best model is termed as the reward. The sorted fitness of all the individuals in the population is treated as the state of the controller. The controller policy $\pi_{\theta_t}$ generates an action $a_t = [m_t, \sigma_t]$, where $m_t$ and $\sigma_t$ be the mean and variance for the sampling of DNN architecture at generation $t$. Given the training dataset $\mathcal{D}^{tr}$ and the validation dataset $\mathcal{D}^{val}$, the algorithmic steps for the GS-EvoN2N is presented in Algorithm \ref{algo:proposed}. Our contributions are highlighted in Algorithm \ref{algo:proposed} and are further discussed in the following sections.
\begin{figure}[!ht]
\centering
\includegraphics[width=7.5cm]{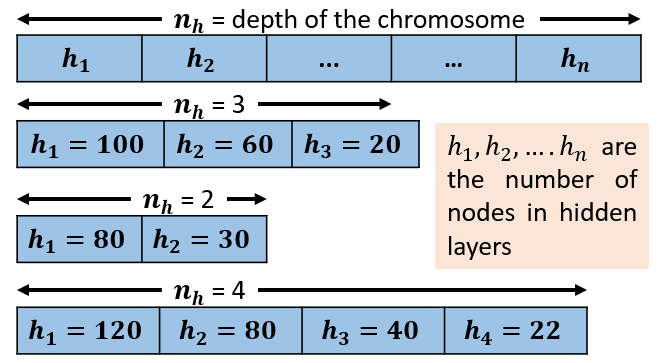}
\caption{Variable-length gene encoding strategy.}
\label{fig:GeneEncoding}
\end{figure}

\begin{algorithm*}
\caption{GS-EvoN2N: The Main Framework}
\label{algo:proposed}
\hspace*{\algorithmicindent} \textbf{Input:} $\mathcal{D}^{tr}\, \&\, \mathcal{D}^{val}$ = training \& validation datasets, $n_R\,\& \,h_R=$ Maximum depth \& width of the DNN respectively\\
\hspace*{\algorithmicindent} \textbf{Output:} $\Psi^{\dagger}$ = best model after the termination or last generation of the evolution.
\begin{algorithmic}[1]
\item $t \longleftarrow 0$ \hspace{3cm}//Set generation count ($t$) = 0;
\item $[m,\;\sigma]\longleftarrow$ Compute mean and variance from allowable range for depth ($n_R$) and width ($h_R$) of the DNN.
\item $\textbf{$P_0$} \longleftarrow \textbf{GuidedPop}(m, \sigma, N_p)$ \hspace{0.3cm} //Generate $N_p$ number of populations using \textbf{Algorithm \ref{algo:popInit}}.
\item $\Psi_0\longleftarrow$ Initialize weight matrices of the first model ($P_0\{1\}$) by small random numbers.
\item $\Lambda,\, \Psi_1^{\dagger}\longleftarrow \textbf{FitnessEval}(P_0, \mathcal{D}^{tr}, \mathcal{D}^{val}, \Psi_0) $  //Evaluate fitness of all individuals in $P_0$ using the \textbf{Algorithm \ref{algo:evafit}}.
\item $\mathcal{R}\longleftarrow NonDominatedSorting(\Lambda_1)$ \hspace{0.3cm} //Assign rank using non-dominated sorting \cite{nsga}.
\item $P_1 \longleftarrow SelectParents(P_0, \; \mathcal{R})\,$ \hspace{1.3cm}//Select parents bybinary tournament selection, \cite{nsga}.
\item ${\Lambda}_{s}\{1\} \longleftarrow \Lambda$ \hspace{3.4cm} //Store the fitness History.
\item $[m,\; \sigma] \longleftarrow \, \textbf{UpdateMeanVar}(P_1,\; {\Lambda}_{s})$ \hspace{1cm} //Update mean and variance term using the \textbf{Algorithm \ref{algo:update_m_sig}}.
\item $Q_1 \longleftarrow \textbf{CrossoverMutation}(P_1, m, \sigma))$ \hspace{0.7cm} //Apply crossover and mutation on $P$ using \textbf{Algorithm \ref{algo:cross}}.
\item $ t\longleftarrow \;t+1$ \hspace{2cm} //Update the generation count
\While{\textit{termination condition is false}}
\item $S_t  \longleftarrow (P_t \cup Q_t)$ \hspace{4.4cm}  //Combine the parent population ($P_t$) \& the child population ($Q_t$).
\item $\Lambda,\, \Psi_{t+1}^{\dagger}\longleftarrow \textbf{FitnessEval}(S_t, \mathcal{D}^{tr},\mathcal{D}^{val}, \Psi_t^{\dagger}) $ \hspace{0.7cm}  //Evaluate fitness of all individuals in $S_t$ using the \textbf{Algorithm \ref{algo:evafit}}.
\item $\mathcal{R}\longleftarrow NonDominatedSorting(\Lambda)$ \hspace{40pt} //Assign rank by non-dominated sorting of fitness $\Lambda$, \cite{nsga}.
\item $\mathcal{K}\longleftarrow CrowdingDistances(S_t, \mathcal{R}, \Lambda)$ \hspace{5pt}  //Find crowding distances of individuals in population set $S_t$, \cite{nsga}.
\item $P_{t+1} \longleftarrow SelectParentsByRankDist(S_t, \mathcal{K}, \Lambda)$ \hspace{26pt} //Select parents by crowding distance and rank, \cite{nsga}.
\item $\Lambda_s\{t+1\} \longleftarrow \Lambda$ \hspace{2.4cm} //Store the fitness History.
\item $[m,\; \sigma] \longleftarrow \, \textbf{UpdateMeanVar}(S_t,\; \Lambda_s)$ \hspace{1cm} //Update mean and variance term using the \textbf{Algorithm \ref{algo:update_m_sig}}.
\item $Q_{t+1} \longleftarrow \textbf{CrossoverMutation}(P_{t+1}, m, \sigma)$  \hspace{15pt} //Apply crossover and mutation on $P$ using \textbf{Algorithm \ref{algo:cross}}.
\If{Termination condition is true}
    \State Exit
\Else
    \State $t \longleftarrow t +1$ \hspace{2.9cm} //Update the generation counter
\EndIf
\EndWhile
\item Return: $\textrm{\textbf{Best Model}}:\;\Psi^{\dagger}\longleftarrow \Psi_{t+1}^{\dagger}$ \hspace{1cm} //Best model of the last generation.
\end{algorithmic}
\end{algorithm*}

\subsection{Population Sampling using Mean and Variance} \label{sec:popInit} AO includes depth and width variation in a defined search space. The real-coded gene encoding strategy is adopted to encode the depth as the number of genes (length of a chromosome) and the number of nodes in a hidden layer as the value of a gene as shown in Fig. \ref{fig:GeneEncoding}.  Let $n_R$ \& $h_R$ be the maximum depth and width of the DNN, then the search space is defined as $[1\; n_R]$ \& $[1\; h_R]$ for depth and the width variations. The mean and variance $m=[m_1, m_2]$ \& $\sigma=[\sigma_1, \sigma_2]$) are initialized as $m_1 = (1+n_R)/2,\;m_2 = (1+h_R)/2$ and $\sigma_1 = (n_R-1)/2, \;\sigma_2 = (h_R-1)/2$.
\begin{algorithm}[!ht]
\caption{GuidedPop: Population Sampling}\label{algo:popInit}
\hspace*{\algorithmicindent} \textbf{Input:} $N$ = Population size, $m=[m_1, m_2]=$ mean \& \\
\hspace*{1.6cm} $\sigma=[\sigma_1, \sigma_2]=$ variance.\\
\hspace*{\algorithmicindent} \textbf{Output:} $P$ = Population with $N$ chromosomes.
\begin{algorithmic}[1]
\item $H\longleftarrow$ generate $N$ Gaussian numbers with $m_1$ and $\sigma_1$.
\For{p = 1 : N}
\item $h\longleftarrow H(p)$ : depth of $p^{th}$ chromosome
\item $tmp\longleftarrow$ generate $h$ Gaussian numbers with $m_2$ and $\sigma_2$.
\item $P\{p\} \longleftarrow $ convert all numbers in $tmp$ to nearest integers.
\EndFor
\item Return $P$
\end{algorithmic}
\end{algorithm}

\subsection{Fitness Evaluation} 
\begin{figure}[!ht]
\centering
\includegraphics[width=8.5cm]{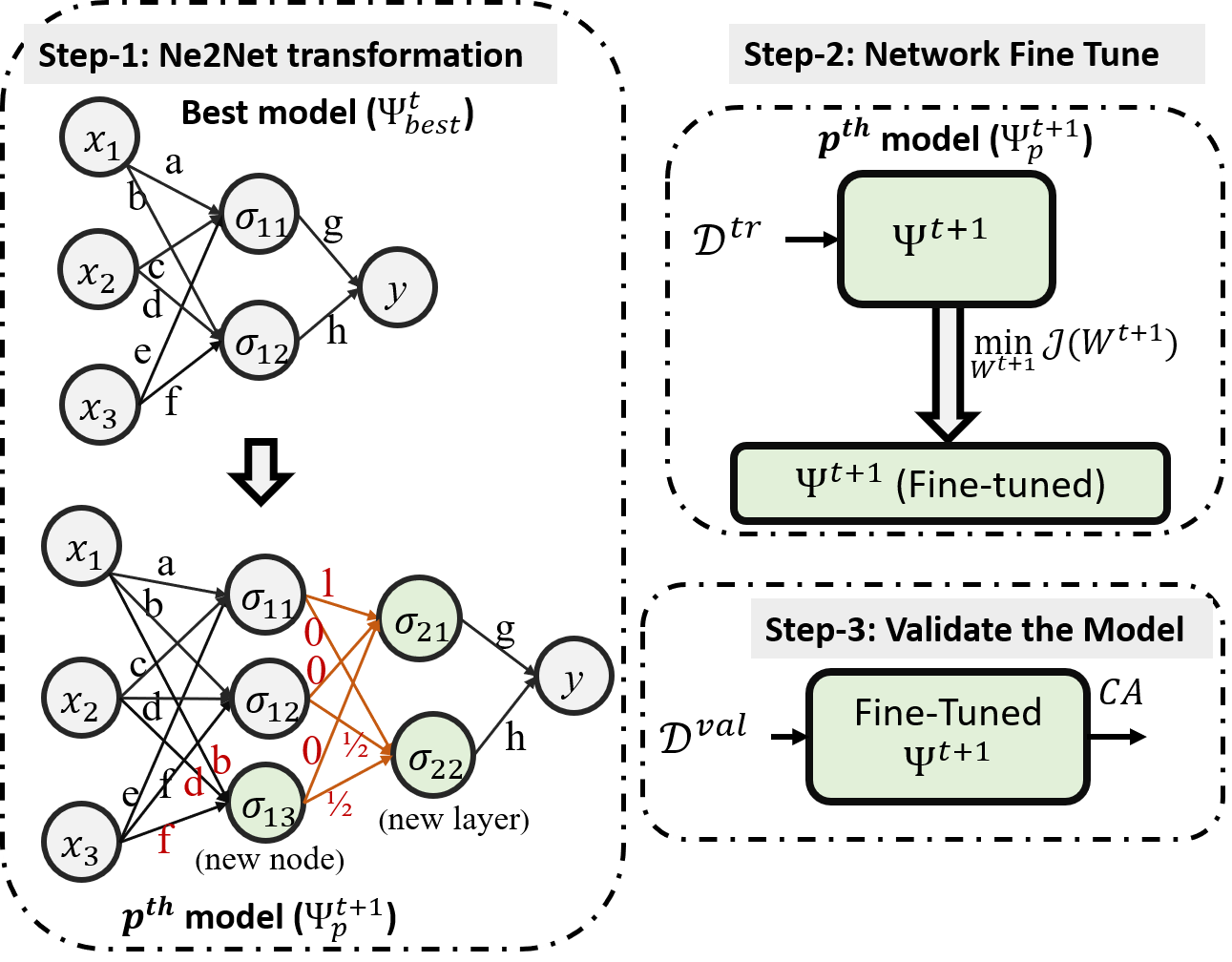}
\caption{Fitness evaluation strategy.}
\label{fig:evafit}
\end{figure}

Fast model evaluation is the most important requirement for NAS, especially when the evolutionary algorithm is used as an AO strategy. If the best model at a generation is transferred for initialization of the DNN weight matrices in the next generation, it makes the training and evaluation of the models faster. The quick learning mechanism suggested in \cite{aks_quick} is adopted for the fitness evaluation as shown in Fig. \ref{fig:evafit}. For the first generation, DNN models are initialized randomly and trained using Limited-Broyden-Fletcher-Goldfarb-Shanno (LBFGS) \cite{bfgs} algorithm. For next-generation and onward, the best model obtained is transformed (Fig. \ref{fig:evafit}) to initialize the models followed by fine-tuning with the LBFGS algorithm for a few iterations only. If a model $\Psi^t$ at generation $t^{th}$ has weight matrix $W^t$, the classification loss $\mathcal{J}$ for a $C$ problem can be defined in term of $[w,\,b]\in W^t$ as
\begin{align}\label{eq:cost}
    \mathcal{J}(W^t) = \frac{1}{n^{s}}\l[\sum_{k=1}^{n^{s}}\sum_{i=1}^{C}I[y_k = c_i]\log{\frac{e^{(w_i^Tf(x_p)+b_i)}}{\sum_{i=1}^C e^{(w_i^Tf(x_p^)+b_i)}}}\r]
\end{align}
where, $f(x)=\Phi(wx+b)$ is the h-level features representation of DNN, $y_k$ be the output label of the $k^{th}$ data sample and $c_i$ denotes the $i^{th}$ class.
The classification accuracy ($CA$)  of fine-tuned model for the validation data is returned as the fitness of that model.
\begin{algorithm}[!ht]
\caption{\textbf{FitnessEval}: Fitness Evaluation}\label{algo:evafit}
\hspace*{\algorithmicindent} \textbf{Input:} $P$ = Population with population size $N_p$,\\
\hspace*{1.6cm} $(\mathcal{D}^{tr}, \mathcal{D}^{val})$ = Training \& validation data.\\
\hspace*{\algorithmicindent} \textbf{Output:} $\Lambda$ = Fitness matrix and $\Psi^{\dagger}$ = Best model.
\begin{algorithmic}[1]
\item $t \longleftarrow$ current generation
\item say $\Psi_t^{\dagger}\longleftarrow \Psi^{\dagger}$ // best model at current generation.
\For{$p \,= \,1 \,: \,N_p$}
\State $\Psi^p_{t}\longleftarrow\textrm{N2N}(\Psi_{t}^{\dagger})$ // Transform $\Psi_{t}^{\dagger}$ to $\Psi^p_t$ $\in$ $P$ using \hspace*{0.5cm}N2N transformation as depicted in step-1 of Fig. \ref{fig:evafit}.
\State Fine-tune the model ($\Psi^p_t$) on $\mathcal{D}^{tr}$ to minimize Eq. (\ref{eq:cost}).
\State $\Lambda(p) \longleftarrow$ Find $CA$ of $\Psi^p_t$ on  dataset $\mathcal{D}^{val}$.
\EndFor
\item $\Psi_t^{\dagger}\longleftarrow \textrm{Best model}$ \hspace{0.1cm} // Find the model with maximum $CA$ and minimum number of model parameters.
\item Return $\Lambda,\; \Psi_t^{\dagger}$
\end{algorithmic}
\end{algorithm}

\subsection{Update Mean and Variance using PG}\label{sec:update_m_s}
Mean ($m$) and variance ($\sigma$) terms are used for sampling of a new population as illustrated in Section-\ref{sec:popInit}. Here, we design a PG-based update laws for $m$ \& $\sigma$ such that the best fitness ($\max(\Lambda)$) is maximized. At any generation $t$, $\max(\Lambda)$ is termed as reward, $a_t=[m, \sigma]$ is termed as action, and weighted average of fitness ($\Lambda$) is termed as state of the policy. Thus, the policy generates $[m^T, \sigma^T]$ to guide the evolution for faster and better convergence. For the design simplification, let us assume that the action is generated by a deterministic policy as 
\begin{equation}
    a_t = f(\theta_t) = \frac{1}{1+e^{-\theta_t}}
\end{equation} 
where, policy parameter $\theta_t$ is selected such that it controls $m\in\Re^{2}$ (mean of depth and width of DNN) and $\sigma\in\Re^{2}$ (variance for depth and width of DNN). The parameter $\theta$ is updated by the policy gradient in (\ref{eq:PG}) calculated using gradient of expected total reward $U_t$ derived in (\ref{eq:PGd}). The total cumulative reward is calculated using fitness matrix $\Lambda_t$ at generation $t$ as in (\ref{eq:reward}). 
\begin{equation}
    \label{eq:reward}
    U_t\;=\; \sum_{i=1}^{t}\frac{\max(\Lambda_i)-\max(\Lambda_{i-1}}{\max(\Lambda_{i-1})}
\end{equation}
The algorithmic steps for the implementation of policy gradient based update of $a_t=[m, \sigma]$ at generation $t$ is summarised in Algorithm \ref{algo:update_m_sig}.

\begin{algorithm}[!ht]
\caption{UpdateMeanVar: Update $m$ \& $\sigma$ using PG}\label{algo:update_m_sig}
\hspace*{\algorithmicindent} \textbf{Input:} $P_t$ = Current population, $\Lambda$ = Fitness matrix,\\
\hspace*{1.6cm} $a_{t-1} = [m, \sigma]$ = Initial mean \& variance term.\\
\hspace*{\algorithmicindent} \textbf{Output:} $a_{t}=[m, \sigma]$ = Updated mean \& variance.
\begin{algorithmic}[1]
\item $N=$ Number of models in $P_t$, $\alpha= $ Learning rate
\item $\bar{n} \longleftarrow$ Compute average depth of models in $P_t$
\item $\Omega =[\omega_p]_{p=1}^{N}$ //Generate a set of weights $\omega_p$ such that $\sum_{p=1}^{N}\omega_p=0$ and $\omega_1>\omega_2>\, ....\, >\omega_{N}$
 \State $\Lambda_t^{sorted}, idx$ = sort($\Lambda_t$, `descending')
 \State $P_t\longleftarrow$ Sort $P_t$ according to $idx$.
\State $n_{p}=$ no. of hidden layers (depth) in $p^{th}$ model.
\State $\delta_p = \max(H_p)-\min(H_p)$ \hspace{0.5cm}//$H_p=$ set of nodes in hidden layers of $p^{th}$ model
 \If{$t\leq 1$}
\State $m=\sum_{j=1}^{N}\l[n_p\omega_p\;\;\; \frac{1}{n_{p}}\sum_{k=1}^{n_{p}}h_{kp}\omega_p\r]$\hspace{0.3cm} //$h_{kp}\in H_p$.
\State $\sigma = \sum_{j=1}^{N}\l[(\bar{n}-n_p)\omega_p\;\;\; \delta_p\omega_p/2\r]$
\Else 
\State $\theta_{t-1} = \ln{\l[a_{t-1}/(1-a_{t-1})\r]}$
\State $U_t\longleftarrow$ Compute cumulative reward using (\ref{eq:reward}).
\State $s_m = \sum_{j=1}^{N}\l[n_p\omega_p \;\;\; \frac{1}{n_{p}}\sum_{k=1}^{n_{p}}h_{kp}\omega_p\r]$
\State $s_{\sigma} = \sum_{j=1}^{N}\l[(\bar{n}-n_p)\omega_p\;\;\; \delta_p\omega_p/2\r]$
\State $s_t = [s_m^T\; s_{\sigma}^T]$
\State $\theta_t\longleftarrow \theta_{t-1} + \alpha*\frac{1}{N}\sum_{k=1}^{N}U_tE[\nabla\log{\pi_{\theta}(s_t|a_{t-1};\theta)}]$
\State $a_t = {1}/{\l(1+e^{-\theta_t}\r)}$
\EndIf
\item Return $a_t$
\end{algorithmic}
\end{algorithm}
\subsection{Crossover and Mutation}
For the optimal search of the network architecture, a combination of exploration and exploitation strategies is adopted. The guided sampling-based generation of new population exploits the search space to force the evolution towards maximum accuracy. To avoid local convergence, $N$ number of individuals are sampled using $m$ and $\sigma$ based on Gaussian distribution, and also $N$ number of parent populations are selected using crowding distance and rank from the current generation.  After that, the two populations are merged to create a double-sized mating pool. Now, the two-step crossover operator introduced in \cite{TL_aks} is applied. The two steps are (i) single point depth crossover ({SPDC}) for depth variation and (ii) common depth simulated binary crossover ({CDSBC}) for gene value (width) alteration. The two-step crossover method is depicted in Fig. \ref{fig:cross}. The whole process of offspring generation is provided in Algorithm \ref{algo:cross}:
\begin{figure}[!ht]
\centering
\includegraphics[width=8.0cm]{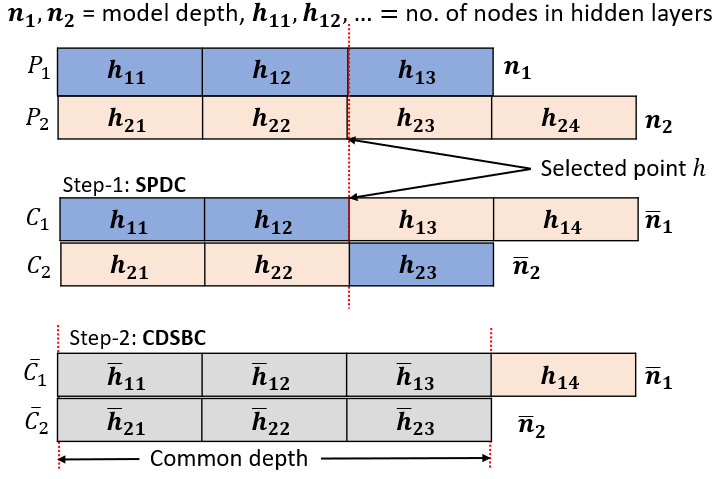}
\caption{Two-steps crossover for chromosomes with different length.}
\label{fig:cross}
\end{figure}
\begin{algorithm}[!ht]
\caption{Offspring Generation: Crossover and Mutation}\label{algo:cross}
\hspace*{0.4cm} \textbf{Input:} $P$ = Parent population, $p_c$ = Crossover probability,\\
\hspace*{1.4cm} $(m, \sigma)$ = mean \& variance term for sampling.\\
\hspace*{0.4cm} \textbf{Output:} $Q$ = Offspring population.
\begin{algorithmic}[1]
\item $Q\longleftarrow$ \textbf{GuidedPop}($m, \sigma, N, n_R, h_R$) //Generate $N$ offspring populations using Algorithm \ref{algo:popInit}.
\item $I_{c}=$ generate random indices of $p_c*100$\% members from $P$.
\For{$i\in I_c$}
\State Select $P_1 = P\{i\} \;\& \;P_2 = Q\{i\}$.
\State Find lengths ($n_1, \;n_2$) of $P_1,\; P_2$ 
\State Set a point $h<\min(n_1, n_2)$ on $P_1,\; P_2$.
\State $C_1,\,C_2\longleftarrow$ \textbf{SPDC} of $P_1,\; P_2$ at point $h$ (as depicted \hspace*{\algorithmicindent}in \textbf{step-1 of Fig. \ref{fig:cross}})
\State $\bar{C}_1, \bar{C}_2 \longleftarrow$ \textbf{CDSBC} for genes of the common depth \hspace*{\algorithmicindent}portion of $C_1,\,C_2$ as depicted in \textbf{step-2 of Fig \ref{fig:cross}}.
\State Replace $Q\{i\}$ by $\bar{C}_1$ or $\bar{C}_2$.
\EndFor
\item Return $Q$
\end{algorithmic}
\end{algorithm}

\section{Experimental Results and Discussion}\label{expD} The efficacy of the proposed framework of GS-EvoN2N is demonstrated on fault diagnosis dataset under different operating conditions taken from (i) Air compressor fault data \cite{sr3}, (ii) Paderborn University (PBU) bearing fault data \cite{paderborn}, and (iii) CWRU bearing fault data \cite{cwru}.

\subsection{Experimental Setup}
\subsubsection{Air Compressor Fault Data \cite{sr3}} \label{sub:air_comp_data} The air compressor data contains acoustic signal recorded on single stage reciprocating type air compressor driven by an $5hp$ induction motor installed at the workshop, EE Department, IIT Kanpur. Data were recorded in eight different cases: healthy and seven different faulty states of the air compressor valve. Therefore, the dataset has 8 classes: (i) Healthy (\textbf{H}), (ii) Leakage Inlet Valve (\textbf{LIV}), (iii) Leakage Outlet Valve (\textbf{LOV}), (iv) Non-Return Valve (\textbf{NRV}), (v) Piston Ring (\textbf{PR}), (vi) Flywheel (\textbf{F}), (vii) Rider Belt (\textbf{RB}), and (viii) Bearing (\textbf{B}). For each class, 225 measurements were taken with 50k samples in each measurement.

\subsubsection{PBU Bearing Fault Data \cite{paderborn}} \label{sub:paderborn_data} PBU bearing fault data is the collection of time-series signals recorded on electrical machine operating under wide variation of shaft load and rotational speed. The four Load and speed settings (LS) are \textbf{LS1: N09\_M07\_F10} (speed = 900 rpm, torque = 0.7 Nm \& radial force = 1000 N), \textbf{LS2:  N15\_M01\_F10} (speed = 1500 rpm, torque = 0.1 Nm \& radial force = 1000 N), \textbf{LS3: N15\_M07\_F04} (speed = 1500 rpm, torque = 0.7 Nm \& radial force = 400 N), and \textbf{LS4: N15\_M07\_F10} (speed = 1500 rpm, torque = 0.7 Nm \& radial force = 1000 N). Total of 32 experimentation with 6 healthy, 12 artificially damaged, and 14 damaged by long run accelerated tests were conducted to record current, vibration signal, radial forces, torque, and bearing temperature. The recorded signals contains two types of faults: inner race \textbf{(IR)} fault and outer race (\textbf{OR}) fault.

\subsubsection{CWRU Bearing Fault Data \cite{cwru}} \label{sub:cwru_data} The CWRU bearing fault data provided by Case Western Reserve University (CWRU) has been a widely used benchmark dataset for bearing fault diagnosis. It contains vibration signal recorded at drive-end (DE) and fan-end (FE) of bearing artificially seeded with inner race fault (\textbf{IR}), outer race (\textbf{OR}), and rolling element ball (\textbf{B}) faults of variable fault diameters (F.D.) (from 0.007 to 0.028 inches). The bearing test rig setup details can be found in \cite{cwru}. 

\subsection{Segmentation and Data Processing} \label{sec:seg}
The recorded time-series signals contain a huge number of samples which is not suitable for training the DNN. To make the dimension of the time-series signals compatible with the DNN, we adopt a segmentation rule with the segment length of approximately $1/4^{th}$ of data points recorded per revolution. Here, we have selected segmentation lengths of 100, 200, \& 400 for the CWRU dataset, Air compressor dataset, and PBU dataset respectively.  
Also, the time-series signals are usually unstructured and not to the scale. Therefore, we have applied the min-max normalization technique to scale down the dataset to $[0,\,1]$. The min-max normalization also removes the effect of outlier points. If for some cases, the outlier points carry some important information, then the z-score minimization technique may be used to make the dataset well-structured \cite{aks_quick}.

\subsection{Dataset Preparation}\label{sec:dataPrep}
For the study of fault diagnosis with the proposed GS-EvoN2N, we prepare the training, the validation, \& the testing dataset under various operating conditions described below.

{\textbf{Case-1 (T1):}} From \textit{Air Compressor Dataset}, $7$ different cases of binary classes and one case of multi-class diagnosis are investigated as listed in table \ref{Table:AirC}. For each class, 4 measurement files (having 50k samples/file) are merged to create a sample of size $1000\times200$ per class taking $200$ points as segment length.

\textit{\textbf{Case-2 (T2):}} From \textit{CWRU FE Dataset}, multi-class diagnosis with class name healthy (\textbf{H}), inner race (\textbf{IR}), outer race (\textbf{OR}), and ball element (\textbf{B}) are considered under different load ($1$, $2$, \& $3$ hp) conditions and different fault diameters (FD) ($7$, $14$, \&  $21$ mil). For each FD (for example, $7$ mil), dataset from all three load conditions are prepared. Thus, the fault diagnosis on total of $9$ cases are presented as listed in table \ref{Table: CWRU}.

\textit{\textbf{Case-3 (T3 \& T4):}} From \textit{PBU Dataset}, two different cases are considered (i) T3: artificially damaged bearing fault and (ii) T4: bearing fault due to long accelerated test. In both cases, multi-class diagnosis with three classes namely \textbf{H-OR-IR} is studied under four load settings (LS) as listed in table \ref{Table: paderborn}. 

Now for the training, the validation, \& the testing, each of the above dataset is split into three portions: $64\%$ train ($\mathcal{D}^{tr}$), $20\%$ test ($\mathcal{D}^{te}$), and $16\%$ validate ($\mathcal{D}^{val}$) datasets.

\begin{table*}[ht]
\centering %
\caption{\textsc{  \small  Air Compressor Dataset (T1): Diagnostic Performance in term of Classification Accuracy}} %
\resizebox{0.74\textheight}{!}{%
\begin{tabular}{ |c|c|c|c|c|c|c|c|c|c|}
\hline
    \multirow{2}{*}{\makecell{Class}}  & \multirow{2}{*}{\makecell{SVM \cite{svm}}} &  \multirow{2}{*}{\makecell{DNN \cite{sparseAE}}}  &  \multirow{2}{*} {DTL \cite{lwen}} &  \multirow{2}{*} {DAFD \cite{wLu}} &  \multicolumn{2}{c|} {N2N \cite{aks_quick}} &  \multirow{2}{*} {EvoDCNN \cite{Ysun2020}} & \multirow{2}{*} {EvoN2N\cite{TL_aks}} & \multirow{2}{*}{\textbf{GS-EvoN2N}} \\
\cline{6-7}
    & & & & & \makecell{W. D. A.} & \makecell{D. A.}  &  &  & \\
\hline
H-LIV & 99.75 & 96.25 & 99.00 & 99.75 & 99.50 & 99.75 & 100.00 & 100.00 & \textbf{100.00} \\
\cline{1-10} H-LOV  &  98.25 & 95.75 & 99.25 & 99.66 & 99.25 & 99.25 & 99.75 & 99.75 & \textbf{100.00} \\  
\cline{1-10} H-PR  &  98.25 & 93.25 & 93.30 & 98.75 & 97.75 & 98.75 & 98.25 & 99.75 & \textbf{99.75} \\  
\cline{1-10} H-B & 98.25 & 98.50 & 98.75 & 98.75 & 96.75 & 98.75 & 98.75 & 99.75 & \textbf{100.00} \\
\cline{1-10} H-F & 99.25 & 99.25 & 99.00 & 98.75 & 99.25 & 99.25 & 99.25 & 100.00 & \textbf{100.00} \\ 
\cline{1-10} H-NRV & 98.75 & 99.00 & 99.00 & 99.75 & 99.00 & 99.25 & 99.25 & 100.00 & \textbf{100.00} \\  
\cline{1-10}
H-RB & 98.25 & 98.25 & 98.25 & 99.00 & 99.75 & 99.75 & 99.25 & 100.00 & \textbf{100.00} \\
\cline{1-10} H-ALL & 97.75 & 99.25 & 99.00 & 99.00 & 99.25 & 99.25 & 99.75 & 99.75 & \textbf{100.00} \\ 
\hline
\rowcolor{Gray}\multicolumn{1}{|c|} {S. D.} & 0.65 & 2.16 & 2.00 & 0.46 & 1.02 & 0.38 & 0.57 & 0.13 & \textbf{0.09} \\ 
\hline
\end{tabular}}
\label{Table:AirC}
\end{table*}
\begin{table*}[ht]
\centering %
\caption{\textsc{  \small  CWRU FE Dataset (T2): Diagnostic Performance in term of Classification Accuracy}} %
\resizebox{0.74\textheight}{!}{%
\begin{tabular}{ |c|c|c|c|c|c|c|c|c|c|c|c|}
\hline
    \multirow{2}{*}{\makecell{Class}}  &
    \multirow{2}{*}{\makecell{FD}}  &  \multirow{2}{*} {\makecell{Load}}  & \multirow{2}{*}{\makecell{SVM \cite{svm}}} &  \multirow{2}{*}{\makecell{DNN \cite{sparseAE}}}  &  \multirow{2}{*} {DTL \cite{lwen}} &  \multirow{2}{*} {DAFD \cite{wLu}} &  \multicolumn{2}{c|} {N2N \cite{aks_quick}} &  \multirow{2}{*} {EvoDCNN \cite{Ysun2020}} & \multirow{2}{*} {EvoN2N\cite{TL_aks}} & \multirow{2}{*}{\textbf{GS-EvoN2N}} \\
\cline{8-9}
    & & & & & & & \makecell{W. D. A.} & \makecell{D. A.}  &  &  & \\
\hline
\multirow{9}{*}{\makecell{H-IR-OR-B}} & \multirow{3}{*}{\makecell{7 mil}} & 1hp & 88.12 & 96.69 & 96.56 & 97.94 & 98.94 & 98.94 & 99.60 & 100.00 & \textbf{100.00} \\
\cline{3-12} && 2hp &  98.12 & 95.94 & 93.44 & 96.12 & 97.12 & 98.12 & 99.60 & 100.00 & \textbf{100.00} \\
\cline{3-12} &&3hp  &  99.10 & 98.75 & 98.75 & 98.44 & 99.44 & 99.44 & 99.70 & 100.00 & \textbf{100.00} \\
\cline{2-12}
& \multirow{3}{*}{\makecell{14 mil}}&  1hp  & 99.10 & 94.75 & 96.88 & 97.19 & 99.19 & 99.67 & 100.00 & 100.00 & \textbf{100.00} \\
\cline{3-12} && 2hp & 98.10 & 95.31 & 92.19 & 95.69 & 97.69 & 98.69 & 98.12 & 99.12 & \textbf{99.60} \\
\cline{3-12} &&3hp & 99.25 & 96.88 & 94.69 & 97.62 & 99.33 & 98.62 & 98.44 & 98.84 & \textbf{100.00} \\
\cline{2-12}
& \multirow{3}{*}{\makecell{21 mil}} &  1hp & 96.88 & 86.56 & 84.69 & 89.62 & 95.62 & 96.62 & 93.75 & 98.75 & \textbf{100.00} \\ 
\cline{3-12} && 2hp & 88.44 & 85.31 & 82.19 & 86.69 & 90.69 & 90.69 & 90.10 & 95.37 & \textbf{98.85} \\
\cline{3-12} && 3hp & 92.19 & 86.56 & 79.38 & 88.06 & 91.06 & 92.06 & 92.81 & 95.81 & \textbf{98.81} \\ 
\hline
\rowcolor{Gray}\multicolumn{3}{|c|} {S. D.} & 4.62 & 5.25 & 7.06 & 4.66 & 3.46 & 3.32 & 3.69 & 1.81 & \textbf{0.51} \\
\hline
\end{tabular}}
\label{Table: CWRU}
\end{table*}

\begin{table*}[ht]
\centering %
\caption{\textsc{\small  CA for Target-2 \& Target-3 Dataset and for very limited samples of Target-2 \& Target-3 Dataset}} %
\resizebox{0.74\textheight}{!}{%
\begin{tabular}{ |c|c|c|c|c|c|c|c|c|c|c|}
\hline
    \multirow{2}{*}{\makecell{Class}}  &  \multirow{2}{*} {\makecell{Data-\\L.S.}}  &  \multirow{2}{*}{\makecell{SVM \cite{svm}}} &  \multirow{2}{*}{\makecell{DNN \cite{sparseAE}}}  &  \multirow{2}{*} {DTL \cite{lwen}} &  \multirow{2}{*} {DAFD \cite{wLu}} &  \multicolumn{2}{c|} {N2N \cite{aks_quick} } &  \multirow{2}{*} {EvoDCNN \cite{Ysun2020}} & \multirow{2}{*} {EvoN2N\cite{TL_aks}} & \multirow{2}{*}{\textbf{GS-EvoN2N}}  \\
\cline{7-8}
    & & & &  & & \makecell{W. D. A.} & \makecell{D. A.} &    &  &  \\
\hline
 \multirow{8}{*}{H-OR-IR} & T3-L1  &  94.25 & 96.92 & 96.92 & 96.92 & 98.64 & 98.94 & 99.25 & 99.75 & \textbf{100.00} \\
\cline{2-11} & T3-L2 &  90.00 & 93.58 & 95.00 & 94.58 & 95.12 & 96.12 & 99.58 & 99.83 & \textbf{99.75} \\        
\cline{2-11} & T3-L3  &  87.17 & 91.92 & 93.33 & 92.08 & 94.44 & 94.44 & 97.50 & 97.70 & \textbf{100.00} \\
\cline{2-11} & T3-L4  &  87.17 & 93.15 & 93.75 & 94.17 & 97.19 & 95.28 & 100.00 & 100.00 & \textbf{100.00} \\
\cline{2-11}
&  T4-L1  &  95.00 & 97.50 & 97.50 & 98.33 & 98.33 & 99.17 & 99.17 & 100.00 & \textbf{100.00} \\
\cline{2-11} & T4-L2  &  92.83 & 95.92 & 96.50 & 96.33 & 96.33 & 96.33 & 98.60 & 99.15 & \textbf{100.00} \\       
\cline{2-11} &T4-L3  &  94.83 & 94.67 & 93.33 & 94.17 & 95.72 & 96.33 & 98.60 & 99.15 & \textbf{100.00} \\
\cline{2-11} & T4-L4  &  94.83 & 95.33 & 95.00 & 95.83 & 95.69 & 95.69 & 93.33 & 98.75 & \textbf{99.75} \\
\hline
\rowcolor{Gray}\multicolumn{2}{|c|} {S. D.} &  3.41 & 1.92 & 1.65 & 1.95 & 1.50 & 1.68 & 2.13 & 0.79 & \textbf{0.10} \\  
 \hline
\end{tabular}}
\label{Table: paderborn}
\end{table*}


\subsection{Implementation Details} For the implementation of the proposed framework of GS-EvoN2N, the initial parameters are selected as: population size ($N$) = 100, crossover probability ($P_c$) = 0.5, and the maximum number of generations is set to very high usually at 50. Also, the termination criteria are set as either the validation accuracy reaches 100\% or it does not change continuously for 3 generations. The allowable ranges for the variation of depth and width are selected as $n_R\in[1, \;10]$ and $h_R\in[10, \;400]$ respectively. The learning rate $\alpha=0.1$. The GS-EvoN2N framework is applied to the training dataset and the best model obtained is tested for the test dataset under all cases (T1, T2, T3, \& T4) described in section \ref{sec:dataPrep}. The classification accuracies ($CA$) are tabulated in tables \ref{Table:AirC}, \ref{Table: CWRU}, \& \ref{Table: paderborn}.

\subsection{Comparison and Analysis} Since our method deals with fault diagnosis with NAS, the major focus has been given on finding the solution for faster selection of the best DNN model for fault diagnosis. Therefore we compare our results with the baseline methods and well as the state-of-the-art methods used for fault diagnosis.  The state-of-the-art methods for intelligent fault diagnosis best reported in various literature are support vector machines (SVM) \cite{svm}, deep neural network (DNN) \cite{sparseAE}, deep transfer learning (DTL) based on sparse autoencoder \cite{lwen}, Deep neural network for domain Adaptation in Fault Diagnosis (DAFD) \cite{wLu}, Net2Net without domain adaptation (N2N\_WDA) \cite{aks_quick}, Net2Net with domain adaptation (N2N\_DA) \cite{aks_quick}, evolutionary deep CNN (EvoDCNN) \cite{Ysun2020}, and evolutionary Net2Net (EvoN2N)\cite{TL_aks}. The DNN, DTL, and DAFD are trained with hidden sizes as $(70-50-20)$. The initial and hyper parameters for EvoDCNN and EvoN2N are kept same as mentioned above. The same dataset (T1, T2, T3, \& T4) are used to train and test all these methods using the procedure suggested in the corresponding cited references. The diagnostic performance in term of $CA$ are tabulated in tables \ref{Table:AirC}, \ref{Table: CWRU}, \& \ref{Table: paderborn}. The standard deviation (S.D.) of $CA$ calculated over the variation in the operating conditions is also tabulated to compare the result deviation with the change in the operating conditions.

\begin{figure}[!ht]
	\centering
	\includegraphics[width=8.7cm]{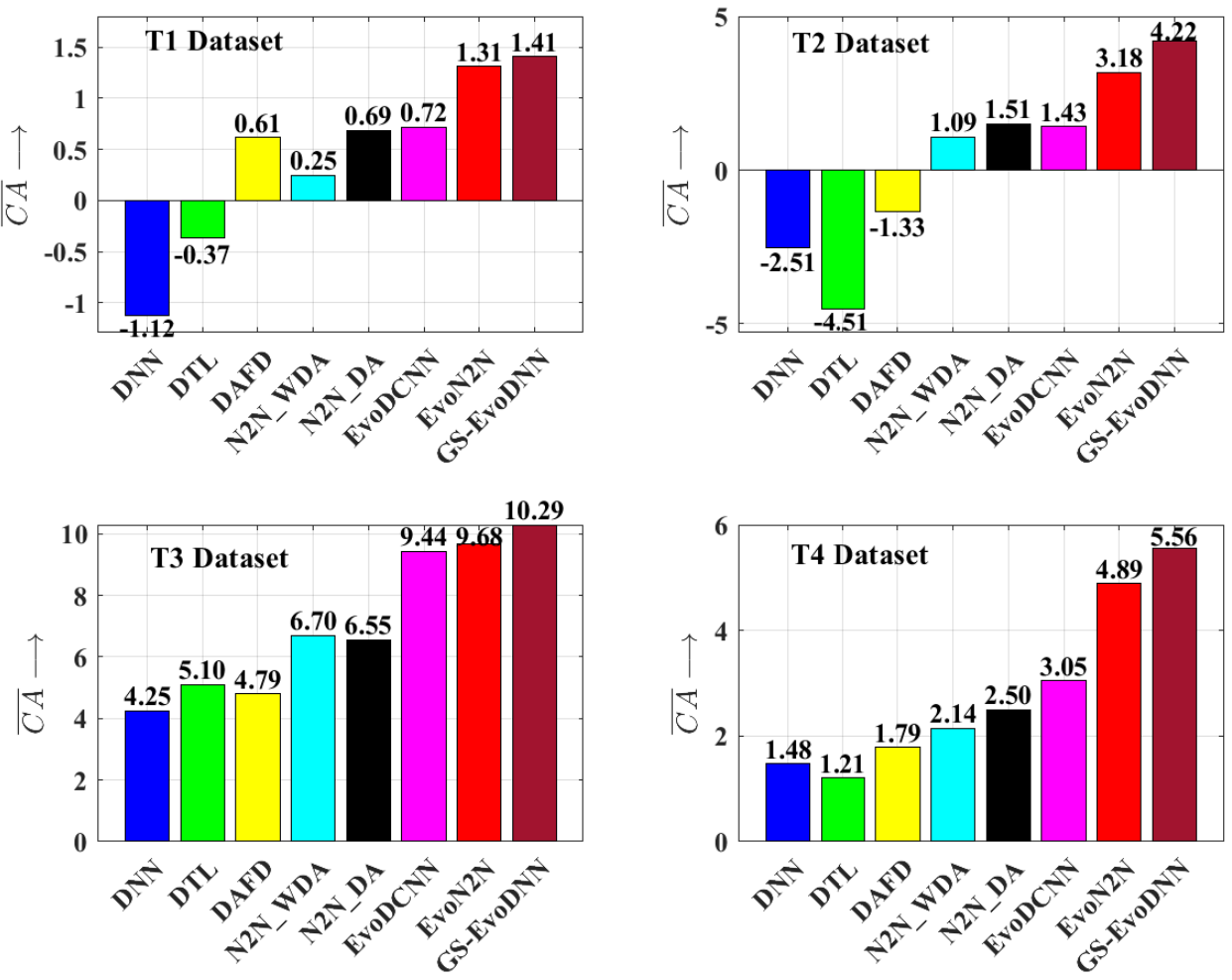}
	\vspace{-5.4cm}
	\caption{TI in term of $\overline{CA}$ for (i) T1: Air Compressor dataset, (ii) T2: CWRU dataset, (iii) T3: PBU dataset with single point fault, and (iv) T4: PBU dataset with distributed fault.}
    \label{fig:TI}
\end{figure}
\begin{figure}[!ht]
	\centering
	\includegraphics[width=8.0cm]{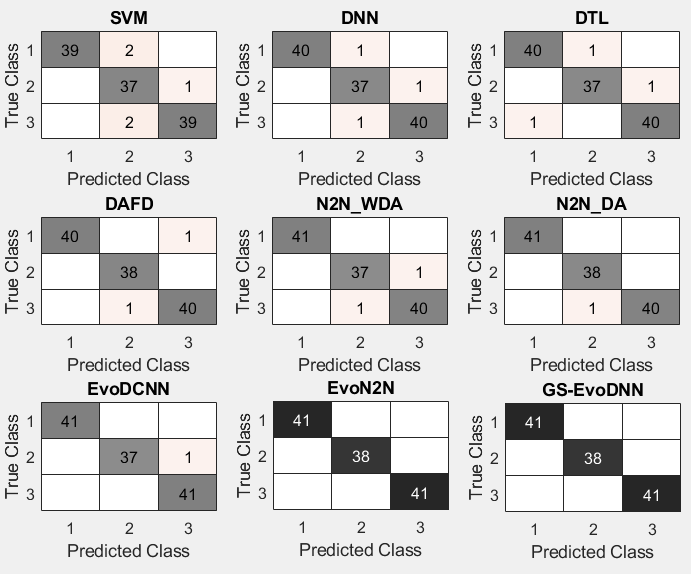}
	\caption{Confusion matrix for dataset T4-L1 (Table \ref{Table: paderborn}): class label \{`1', `2', `3'\} represents the class name \{`H', `OR', `IR'\}.}
    \label{fig:cm}
\end{figure}
\begin{figure}[!ht]
	\centering
	\includegraphics[width=8.0cm]{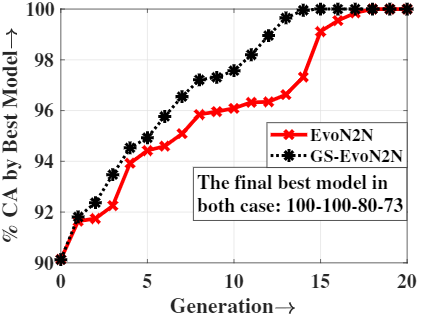}
	\caption{Rise of $CA$ curve of GS-EvoN2N and EvoN2N with generation (CWRU FE, 7 mil FD, 1 hp load).}
    \label{fig:acc_curve}
\end{figure}

\subsection{Discussion} The diagnostic performances by the proposed method and the selected state-of-the-art methods conclude the following observations.
\begin{enumerate}
    \item The $CA$ comparison in tables \ref{Table:AirC}, \ref{Table: CWRU}, \& \ref{Table: paderborn} reveal that the diagnostic performances are very much affected by the model selection. The DNN model with the best suitable architecture for the given dataset can perform up to almost 100\% $CA$ while other methods with pre-selected architecture fail to perform well. 
    \item Considering SVM \cite{svm} as the baseline diagnostic method, We evaluate the transfer improvement ($TI$) in term of average $CA$ for the dataset T1, T2, T3, \& T4 separately. If the average $CA$ is denoted as $\overline{CA}$, the $TI$ is defined as $TI = \overline{CA} - \overline{CA}_{b}$, where $\overline{CA}_{b}$ is the average $CA$ by SVM. The $TI$ graph shown in Fig. \ref{fig:TI} shows the performance improvement of the proposed framework in comparison with the state-of-the-art methods and the baseline method `SVM'. 
    \item Fig. \ref{fig:cm} demonstrate the classification performance by confusion chart matrices for one of the dataset (T4-L1: table \ref{Table: paderborn}). The confusion matrices with blackened diagonal elements represent the classifications with 100\% accuracies. The confusion matrices with grey shades are the methods with missed classifications. All the test samples are classified correctly by the proposed method 'GS-EvoN2N', therefore the proposed method is capable of selecting the best possible architecture that has almost 100\% diagnostic performance.
    \item Fig. \ref{fig:acc_curve} show the evolution of the best model for the proposed GS-EvoN2N and the EvoN2N \cite{TL_aks} with the generation. The comparison of the rise of the curve reveals that guided sampling helps the algorithm to attain optimality faster. Therefore, the proposed method GS-EvoN2N requires less number of generations to converge to the global optima.
    \item The comparison between EvoDCNN \cite{Ysun2020}, EvoN2N \cite{TL_aks}, \& the proposed GS-EvoN2N reveal that the fully connected (DNN)  model with the best architecture is more suitable for fault diagnosis applications compared to the CNN model in EvoDCNN \cite{Ysun2020} proposed for image classification.
\end{enumerate}

\subsection{Complexity Analysis} The worst complexities in one iteration of the entire algorithm \ref{algo:proposed} are contributed by (i) fitness evaluation of the DNN model and (ii) the non-dominated sorting. The complexity of the fitness evaluation of DNN is mainly contributed by parameter optimization by L-BFGS which is $O(N_I*n^2)$, where $n\, \&\, N_I$ be the total number of parameters and number of iterations required to fine-tune the DNN model. The non-dominated sorting algorithm has a complexity of $O(MN_p^2)$, where $M \,\&\, N_p^2$ be the number of objectives and the population size respectively. Since $M$ is very small compared to $N_I$, therefore, the time complexity for GS-EvoN2N with population size $N_p$ is given by $O(N_IN_pn^2)$, where $n=$ total number of parameters in one model and $N_I$ = number of iterations taken for model training.

\section{Conclusions} In this article, we have formulated a guided sampling-based evolutionary algorithm for DNN architecture search. The proposed framework uses the concept of policy gradient to sample the new population to force the evolution towards the maximization of classification performance. The classification accuracies of the best model at each generation are used as a reward to update the policy parameters. The policy controller generates the mean and variance term to sample the new architecture for better performance. The best model obtained at each generation is also transferred to the next generation to initialize the model evaluation using the concept of net2net transformation. The entire algorithm becomes faster to attain the global maxima. Therefore, this method is very good in terms of faster evolution and faster convergence while ensuring global convergence. The validation using dataset under various cases from Air Compressor data, CWRU data, and PBU data prove that the proposed framework is capable to obtain the best model to get diagnostic performance almost up to 100\% accuracy. This method can also be employed for the architecture optimization of the CNN model with image classification or object detection applications.
\label{conclusion} 

\ifCLASSOPTIONcaptionsoff
  \newpage
\fi

\bibliographystyle{IEEEtran.bst}
\bibliography{Reference.bib}

\end{document}